\documentclass[11pt,a4paper]{article}
\usepackage{authblk}
\usepackage[hyperref]{acl2021}
\usepackage{times}
\usepackage{latexsym}

\usepackage{microtype}
\usepackage{graphicx}
\usepackage{tabu}
\usepackage{xcolor}
\usepackage{float}
\usepackage{lipsum,multicol}
\usepackage{caption, subcaption}
\usepackage{multirow}
\usepackage{blindtext}
\usepackage{stfloats}
\usepackage{colortbl}
\usepackage{hyperref}

\makeatletter
\newcommand{\printfnsymbol}[1]{%
  \textsuperscript{\@fnsymbol{#1}}%
}
\makeatother

\aclfinalcopy 
\def\aclpaperid{2035} 

\title{Constructing Multi-Modal Dialogue Dataset by\\Replacing Text with Semantically Relevant Images}
\author{\textbf{Nyoungwoo Lee\thanks{\hspace{0.2cm} Equal contribution.} }}
\author{\textbf{Suwon Shin\printfnsymbol{1}}}
\author{\textbf{Jaegul Choo}}
\author{\textbf{Ho-Jin Choi}}
\author{\textbf{Sung-Hyon Myaeng}}
\affil{KAIST, Daejeon, South Korea}
\affil{\textit {\{leenw2, ssw0093, jchoo, hojinc, myaeng\}@kaist.ac.kr}}

\begin{document}
\maketitle
\begin{abstract}
In multi-modal dialogue systems, it is important to allow the use of images as part of a multi-turn conversation. Training such dialogue systems generally requires a large-scale dataset consisting of multi-turn dialogues that involve images, but such datasets rarely exist. 
In response, this paper proposes a 45k multi-modal dialogue dataset created with minimal human intervention. Our method to create such a dataset consists of (1) preparing and pre-processing text dialogue datasets, (2) creating image-mixed dialogues by using a text-to-image replacement technique, and (3) employing a contextual-similarity-based filtering step to ensure the contextual coherence of the dataset.
To evaluate the validity of our dataset, we devise a simple retrieval model for dialogue sentence prediction tasks. Automatic metrics and human evaluation results on such tasks show that our dataset can be effectively used as training data for multi-modal dialogue systems which require an understanding of images and text in a context-aware manner.
Our dataset and generation code is available at \url{https://github.com/shh1574/multi-modal-dialogue-dataset}.
\end{abstract}

\section{Introduction}
\label{introduction}

Humans often use images in instant messaging services to express their meaning and intent in the dialogue context. For a dialogue system such as a chatbot to respond to human users adequately in this kind of multi-modal situations, it is necessary to understand both images and texts in their context and incorporate them in the dialogue generation process.

Training such a multi-modal dialogue system generally requires a large amount of training data involving images and texts in various contexts. However, numerous existing approaches relying on image captioning~\citep{lin2014microsoft, young2014image} or visual question answering~\citep{mostafazadeh2016generating, das2017visual} techniques had to be trained with the datasets mostly irrelevant to the dialogue context. In other words, images were interpreted independently of the dialogue context, due to the lack of sufficient multi-modal dialogue datasets.

\begin{figure}[t]
\centering
\begin{tabular}{c}
     \includegraphics[width=0.47\textwidth]{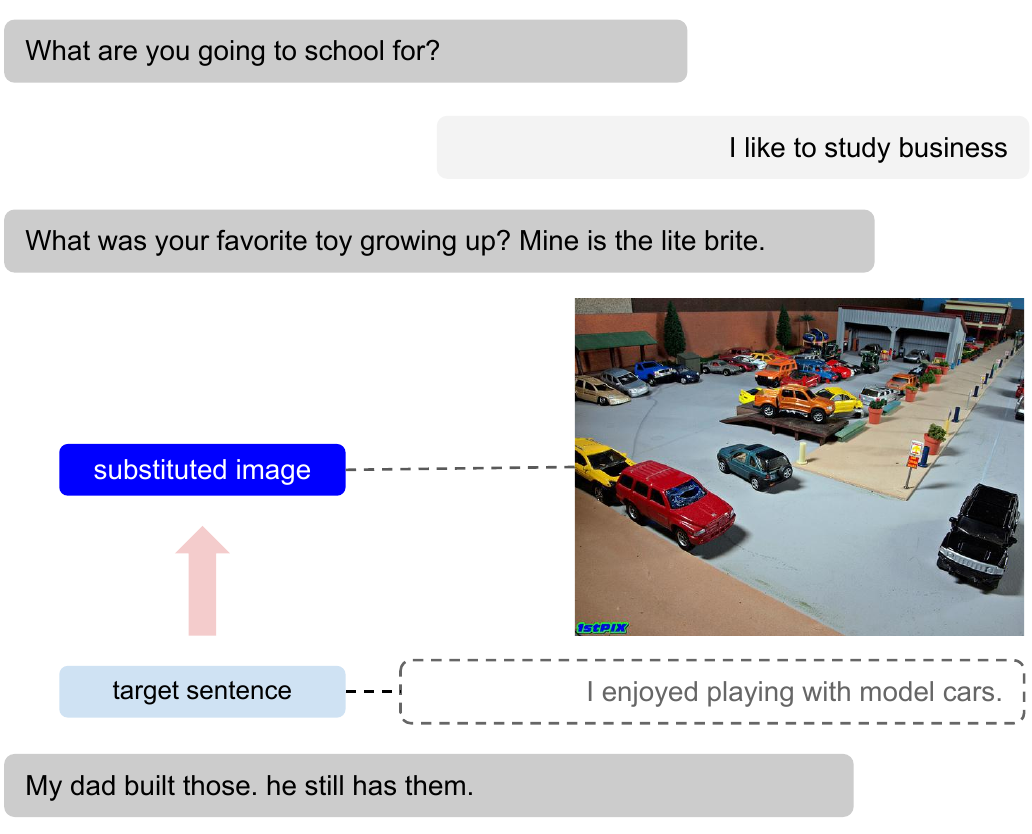}
\end{tabular}
\caption{Example of multi-modal dialogue dataset.}
\label{fig:multimodaldialogue}
\end{figure}

Those datasets containing image-grounded conversations~\citep{mostafazadeh2017image, shuster2020image} do not even cover the situations related to dialogue context before the image, because all conversations in the dataset always start from the given image. Although the relationship between images and texts can be learned using image-grounded conversations~\citep{lu2019vilbert, chen2019uniter, tan2019lxmert, su2019vl, li2019visualbert}, it cannot still learn the dependency between the dialogue context before and after the image.

In this paper, we propose a 45k multi-modal dialogue dataset in the form of Fig.~\ref{fig:multimodaldialogue}.
Each multi-modal dialogue instance consists of a textual response and a dialogue context with multiple text utterances and an image. To create this dataset, we start with existing text-only dialogue datasets as source dialogues, and then replace part of sentences in source dialogues with their semantically relevant images. The detailed steps include (1) source dialogue pre-processing, such as deleting a stop word, to improve the quality of similarity calculations, (2) creating dialogues containing an image by replacing a sentence with a similarity-based text-to-image replacement technique, and (3) pruning low-quality dialogues by employing a contextual-similarity-based filtering method. 
The overall process ensures that the created dataset consists of natural dialogue examples containing diverse images.

\begin{figure*}[ht]
\centering
\begin{tabular}{c}
     \includegraphics[width=1\textwidth]{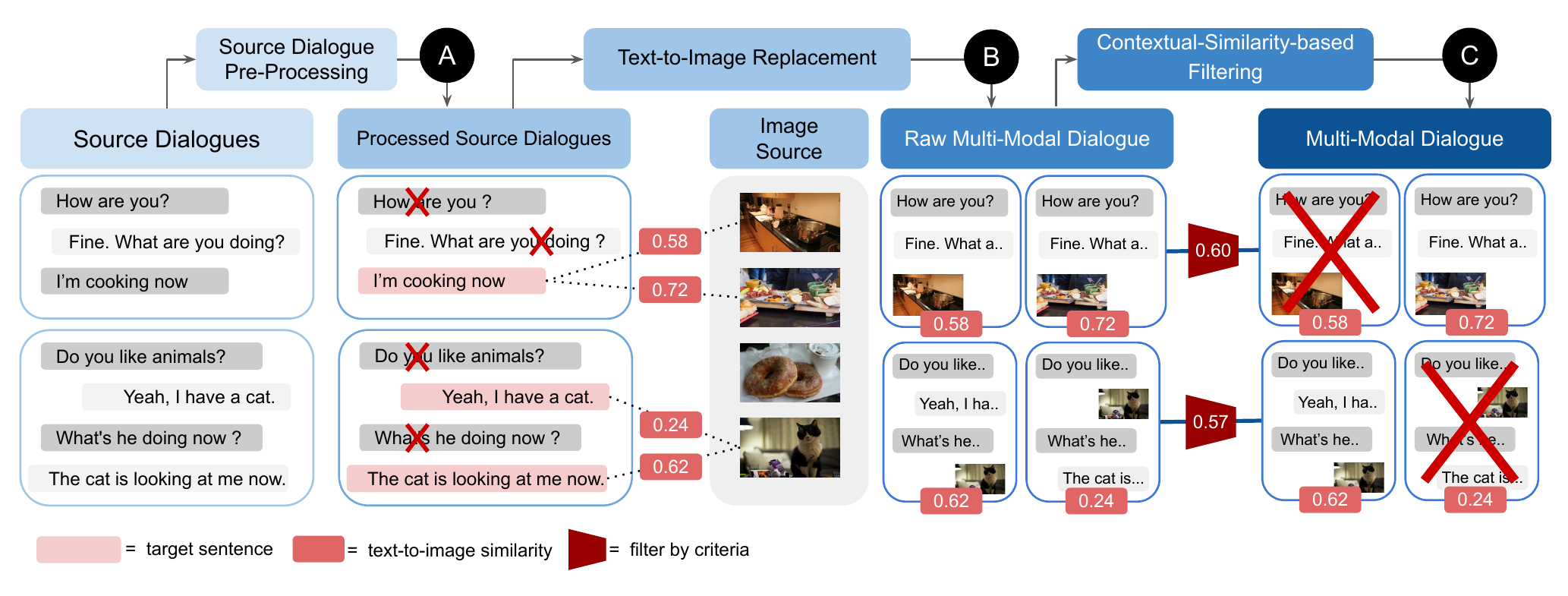}
\end{tabular}
\caption{Overall pipeline for multi-modal dialogue dataset creation.}
\label{fig:creationprocess}
\end{figure*}

In order to validate our dataset creation process and examine the quality of our multi-modal dialogue dataset, we devise the task of predicting current and next dialogue sentences while considering the dialogue context and images. We also develop simple retrieval models to learn the relationship between images and texts for the tasks.
Human evaluation results for predicting dialogue tasks show that the sentences are predicted as intended, i.e., in a context-aware manner, using the images. The results also show that our dataset can serve as practical training resources for multi-modal dialogue tasks that involve both image and dialogue context.

\section{Multi-Modal Dialogue Generation}
\label{datacreation}
Our multi-modal dialogue dataset is constructed based on three source dialogue datasets and two image captioning datasets: DailyDialog~\citep{li2017dailydialog}, EmpatheticDialogues~\citep{rashkin2018towards}, and Persona-Chat~\citep{zhang2018personalizing} for the former and the MS-COCO~\citep{lin2014microsoft} and Flicker 30k~\citep{young2014image} for the latter. The statistics of each dataset are summarized in Appendix~\ref{sourcedatastatistic}. After obtaining the source datasets, we replace sentences in the source dialogues with proper images by searching the image dataset to create image-mixed dialogues that maintain semantic coherence. To this end, we apply the three-stage method as shown in Fig.~\ref{fig:creationprocess}: (1) source dialogue pre-processing, (2) text-to-image replacement, and (3) contextual-similarity-based filtering.

\paragraph{Source Dialogue Pre-Processing}
We pre-process source dialogue datasets for the subsequent text-to-image replacement (A in Fig.~\ref{fig:creationprocess}). To select candidate dialogue sentences to be replaced by images, we first exclude the question sentences from the candidate dialogues because it is not realistic to infer back a question out of an image to put in the place of the question. This step filters out 25.08\% of the total sentences in the source dialogue datasets.
Second, we remove stop words from the source dialogue datasets, because they do not contain meaningful information. All the remaining sentences in dialogue contexts after the pre-processing step are considered as potential target sentences to replace.

\paragraph{Text-to-Image Replacement}
In this step, we create multi-modal dialogues containing images by replacing target sentences from the candidate dialogue sentences with appropriate images in the image dataset based on text-to-image similarity (B in Fig.~\ref{fig:creationprocess}).
We calculate the similarity by the pre-trained Visual Semantic Reasoning Network (VSRN)~\citep{li2019visual}, a state-of-the-art image-text matching model based on text-to-image similarity.
We first identify target sentences and then select candidate images for replacement using the threshold ensuring context coherence, as will be discussed in the subsequent contextual-similarity-based filtering step.
Because we aim to maintain the comprehensive flow of the dialogue, we replace only one sentence with an image per dialogue. 
If multiple image candidates exist for a single sentence, we separate them into distinct image-mixed dialogue instances. In detail, such separated instances are all made up of the same dialogue context and text response except for substituted images.

\paragraph{Contextual-Similarity-based Filtering}
\label{filtering}
We employ a contextual-similarity-based filtering step to enhance the context coherence of the created image-mixed dialogues (C in Fig.~\ref{fig:creationprocess}). We filter out the dialogues where text-to-image similarity does not exceed the threshold determined by human annotators. For human annotators on the matching quality of an image, a total of 300 test dialogues are selected for each combination. Since we used three source dialogue datasets and two image datasets, we create six combinations of each dialogue dataset and each image dataset. Automatically created image-mixed dialogue instances are divided into ten segments based on the similarity values, and 30 are selected randomly from each. We hired a total of 18 annotators to evaluate 1,800 instances sampled from these six combinations. The evaluation system is described in Appendix~\ref{humanevalsys}.

The human evaluation was conducted based on three questions for each instance:
\begin{itemize}
\item Q1: How well does the substituted image contain \textbf{key objects} in the target sentence?
\item Q2: How well does the substituted image represent the \textbf{meaning} in the target sentence?
\item Q3: When the image is substituted for the target sentence, how \textbf{consistent} is it with the \textbf{context} of the conversation?
\end{itemize}
Q1 and Q2 ask whether the substituted image contains the core meaning of the target sentence (on a 3-point scale). Q3 evaluates the context coherence of the created dialogue containing the image (on a 5-point scale). 
We assume that dialogues above the median of the evaluation score (2 for Q1, Q2, and 3 for Q3) are suitable for use as training instances. Based on this assumption, we determine the threshold for each combination by interpolating the median in the correlation graph of the evaluation results and the similarity (Appendix~\ref{humanbase}).
We then analyze the correlation between the score for each question and text-to-image similarity using Spearman's correlation analysis as shown in Table~\ref{table:spearman}. Overall, the similarity values are positively correlated with the scores obtained for the questions.
Since Q2 and Q3 are reasonably correlated with semantic similarity, the substituted images tend to reflect the meaning of the target and context sentences. Thus, the evaluation results indicate that the automatically created image-text pairs with high similarity can be used as multi-modal dialogues. We filter the generated multi-modal dialogues based on the determined similarities, and then set the filtered dialogues as our final dataset. The statistics of the final dataset are summarized in Table~\ref{table:final_statistics}.

\begin{table}[t]
\centering
    {\small
    {\tabulinesep=0.6mm
    \begin{tabular}{l|cccc}
    \hline 
    & Similarity & Q1 & Q2 & Q3 \\
    \hline
    \\[-1em]
    Similarity & \cellcolor{lightgray} & \textbf{0.5893} & 0.4422 & 0.4334 \\
    Q1 & \cellcolor{lightgray} & \cellcolor{lightgray} & 0.7103 & 0.6646 \\
    Q2 & \cellcolor{lightgray} & \cellcolor{lightgray} & \cellcolor{lightgray} & \textbf{0.7570} \\
    Q3 & \cellcolor{lightgray} & \cellcolor{lightgray} & \cellcolor{lightgray} & \cellcolor{lightgray}\\
    \hline
    \end{tabular}
    }}
\caption{\label{table:spearman}Spearman's correlation ${\rho}$ between three questions and text-to-image similarity.}
\end{table}

\begin{table}[t]
\centering
    {\small
    {\tabulinesep=0.6mm
    \begin{tabu}{lccc}
    \hline 
    & train & valid & test \\
    \hline
    \# total dataset & \textbf{39956} & \textbf{2401} & \textbf{2673} \\
    Avg length of dialogue turns & 13.01 & 13.62 & 13.59 \\
    Avg length of sentences & 51.47 & 50.76 & 50.70 \\
    \# total unique images & 12272 & 334 & 682 \\
    \# total unique dialogues & 13141 & 2148 & 2390 \\
    \# total unique target sentences & 21495 & 2400 & 2671 \\
    \scriptsize{Avg \# of substituted images in a dialogue} & 1.86 & 1.00 & 1.00 \\
    Avg \# of targets in a dialogue & 1.64 & 1.12 & 1.12 \\
    \hline
    \end{tabu}
    }}
\caption{\label{table:final_statistics}Multi-modal dialogue dataset statistics for splits of training, validation, and test set.}
\end{table}

\paragraph{Data Quality}
We evaluate the quality of our dataset to validate the proposed dataset creation method. To this end, we randomly sample 300 image-mixed dialogues from our final dataset. The evaluation proceeds in the same manner as before, but we add a new question Q4, which asks to choose the intent of the image used in the dialogue as one among (1) answering the question, (2) expressing emotional reactions, (3) proposing a new topic, and (4) giving additional explanations for the previous context. For Q1, Q2, and Q3, the average scores evaluated by three annotators are shown to be 2.56, 2.17, and 3.13, respectively, indicating that the context of the conversation containing the substituted image is consistent in our dataset. For Q4, the responses from the annotators are distributed with 27.3\%, 20.0\%, 32.7\%, and 14.7\%, for the four intent types as mentioned above, indicating our dataset contains balanced intent types.

\section{Experiments}
\label{experiments}

\subsection{Experimental Setup}

We consider two dialogue sentence prediction tasks given an image and a dialogue: current dialogue prediction and next dialogue prediction for a given image.
We use a simple retrieval model composed of three modules \citep{shuster2020image, shuster2020multi}: Resnext-101~\citep{xie2017aggregated} for an image encoder, BERT~\citep{devlin2018bert} for a text encoder, and the fusion module. As input for training the model, we use images and up to three dialogue sentences immediately before the images as dialogue context.

\begin{table}[t]
\centering
    {\small
    {\tabulinesep=0.6mm
    \begin{tabu}{ll|ccc}
    Model & Task & R@1 & R@5 & \scriptsize{Mean Rank}\\
    \hline
    IR Baseline & Current & 21.62 & 49.49 & 30.04\\
    IR Baseline & Next & 8.13 & 21.07 & 29.41\\
    \hline
    Retrieval Model & Current & \textbf{50.35} & \textbf{86.64} & \textbf{3.11}\\
    Retrieval Model & Next & 14.38 & 36.10 & 20.58\\
    \end{tabu}
    }}
\caption{\label{table:summary_auto}Automatic evaluation results about retrieval models and an information retrieval baseline on the current and next dialogue prediction task.}
\end{table}

\subsection{Automatic Evaluation}
We perform quantitative comparisons that follow recent work~\citep{shuster2020image} to find the optimal setting for our retrieval model (Appendix~\ref{bestmodel}).
To evaluate the retrieval accuracy, we use the recall at 1 and 5 out of 100 candidates consisting of 99 candidates randomly chosen from the test set and 1 ground-truth sentence, called R@1/100 and R@5/100, respectively. We also use the mean reciprocal rank.
We compare our model with a simple information retrieval baseline. The candidates of the baseline model are ranked according to their weighted word overlap between the target sentence and an image caption followed by dialogue context.

As shown in Table~\ref{table:summary_auto}, the R@1 performance of the retrieval model obtained 50.35 and 14.38 on the current and next sentence prediction task, outperforming the baseline on both tasks.
This result indicates that our dataset properly works as the training data to learn the relationship between images and dialogue context in dialogue sentence prediction tasks where images and dialogue context have to be considered together.

\begin{table}[t]
\centering
    {\small
    {\tabulinesep=0.6mm
    \begin{tabu}{l|ccc}
    Model inputs & R@1 & R@5 & \scriptsize{Mean Rank}\\
    \hline
    Image Only & 37.30 & 80.66 & 3.91\\
    Dialogue Context Only & 28.06 & 56.83 & 12.57\\
    Image + Dialogue Context & \textbf{51.21} & \textbf{86.34} & \textbf{3.08}\\
    \end{tabu}
    }}
\caption{\label{table:ablation1}Ablation studies of our retrieval models on the current dialogue prediction task.}
\end{table}

\begin{table}[t]
\centering
    {\small
    {\tabulinesep=0.6mm
    \begin{tabu}{l|ccc}
    Model inputs & R@1 & R@5 & \scriptsize{Mean Rank}\\
    \hline
    Image Only & 7.29 & 21.92 & 31.78\\
    Dialogue Context Only & 11.90 & 29.89 & 23.95\\
    Image + Dialogue Context & \textbf{14.38} & \textbf{36.10} & \textbf{20.58}\\
    \end{tabu}
    }}
\caption{\label{table:ablation2}Ablation studies about our retrieval models on the next dialogue prediction task.}
\end{table}

\subsection{Ablation Study}
We then conduct ablation studies by removing modalities (image and dialogue context) in turn to check whether unwanted correlations exist in our dataset.
Since we created our training and test datasets by a semi-automatic data creation method, unwanted correlations can exist in datasets that can infer the correct answer without using the image and context simultaneously. Such correlations would prevent the model from properly learning the relationship between images and context.

As shown in Tables~\ref{table:ablation1} and~\ref{table:ablation2}, the results first show that the recall measure for ground-truth answers in the model that considers both context and image is higher than the model considering only images. It indicates that the models in each task properly consider both images and dialogue context to predict sentences. To elaborate, the model that only considers images are likely to choose responses that do not match the dialogue context before the image. For example in a given dog photo shown during a sad mood conversation, the model that only considers images can generate an out-of-context response, such as ``It is so cute.''. On the other hand, in the same context, the model that considers both the context and the image could generate appropriate responses, such as ``what is wrong with your dog?'' or ``I miss your dog.''.

The overall tendency also shows that the model performance degrades when we delete each modality one by one. Such results suggest that our data creation process did not generate correlations that interfere with forming the relationship between images and dialogue context.

\subsection{Human Evaluation}
We create a new test set to confirm that the model can predict sentences well even on test dialogues that are not constructed in the same manner. To this end, two researchers manually created 100 multi-modal dialogues by adding images to source dialogues that were not used in our dataset generation process for human evaluation.
We proceed with the evaluation with three annotators per each prediction task, using a question (on a 5-point scale) asking how much the sentences predicted by the model are relevant to the image and dialogue context.
The average scores of three annotators for each task were shown to be 3.36 for the current turn prediction and 3.06 for the next turn prediction.
The results indicate that the models can predict sentences in a context-aware manner even with dialogues organized by humans.

\section{Conclusions}
\label{conclusion}
We present the multi-modal dialogue dataset consisting of 45k multi-turn dialogues containing semantically coherent images as well as the dataset creation method.
Human evaluation results of our multi-modal dialogues reveal that context coherence is well maintained even if the sentence is replaced by an image, showing the validity of our dataset and data creation approach.
We then evaluate our dataset using two multi-modal dialogue prediction tasks, demonstrating its effectiveness when training a dialogue system to learn the relationship between images and dialogue contexts.
Our proposed data creation method can be applied when efficiently preparing large-scale multi-modal dialogue datasets that cover diverse multi-modal situations.

\section*{Acknowledgments}
This work was supported by Institute for Information \& communications Technology Planning \& Evaluation(IITP) grant funded by the Korea government(MSIT) (No. 2013-2-00131, Development of Knowledge Evolutionary WiseQA Platform Technology for Human Knowledge Augmented Services, No. 2019-0-00075, Artificial Intelligence Graduate School Program(KAIST), and No. 2020-0-00368, A Neural-Symbolic Model for Knowledge Acquisition and Inference Techniques)

\bibliographystyle{acl_natbib}
\bibliography{anthology,acl2021}

\appendix
\clearpage
\section{Source Datasets Statistics}
\label{sourcedatastatistic}

\begin{table}[h]
\centering
    {\small
    {\tabulinesep=0.6mm
    \begin{tabu}{lcccc}
    \hline 
    & type & training & validation & test \\
    \hline
    DailyDialogue & dialog & 11118 & 1000 & 1000 \\
    Persona-Chat & dialog & 8938 & 999 & 967 \\
    \scriptsize{EmpatheticDialogues} & dialog & 17792 & 2758 & 2539 \\
    \hline
    MS-COCO & image & 113287 & 5000 & 5000 \\
    Flickr 30k & image & 28000 & 1000 & 1000 \\
    \hline
    \end{tabu}
    }}
\caption{\label{table:source_statistics}Source dialogue and image captioning dataset statistics for splits of training, validation, and test set.}
\end{table}

\section{Detailed Description of Contextual-Similarity-based Filtering}
\label{humanbase}

\begin{figure*}[!t]
\centering
\begin{tabular}{ccc}
     \includegraphics[width=0.32\textwidth]{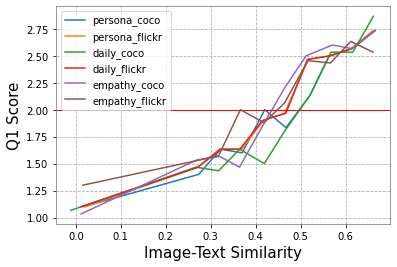}
     \includegraphics[width=0.32\textwidth]{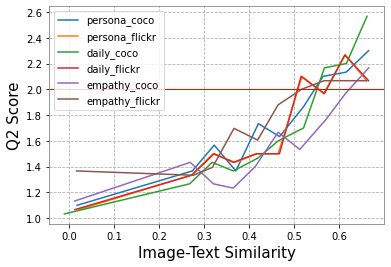}
     \includegraphics[width=0.32\textwidth]{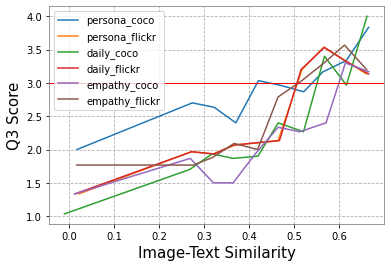}
\end{tabular}
\caption{Correlation between text-to-image similarity and question scores (Q1, Q2, and Q3) for six combinations.}
\label{fig:threshold_result}
\end{figure*}

\begin{table}[h]
\centering
    {\small
    {\tabulinesep=0.6mm
    \begin{tabu}{lcccc}
    \hline 
    & \scriptsize{threshold} & train & valid & test \\
    \hline
    Persona-COCO & 0.546 & 11606 & 411 & 1136 \\
    Persona-Flickr & 0.509 & 19148 & 1654 & 1014 \\
    Daily-COCO & 0.555 & 3418 & 47 & 319 \\
    Daily-Flickr & 0.619 & 141 & 6 & 5 \\
    Empathetic-COCO & 0.623 & 245 & 2 & 11 \\
    Empathetic-Flickr & 0.516 & 5398 & 281 & 188 \\
    \textbf{Total} & & \textbf{39956} & \textbf{2401} & \textbf{2673} \\
    \hline
    \end{tabu}
    }}
\caption{\label{table:filter_by_thres}Number of data instances filtered by the thresholds for each combination}
\end{table}

In this section, we analyze the human evaluation results for contextual-similarity-based filtering and determine thresholds for each dataset combination. The correlations between the similarity and evaluation results for each question are shown in Fig.~\ref{fig:threshold_result}. We assume that dialogue instances above the median of the evaluation score (2 for Q1, Q2, and 3 for Q3) are suitable for use in training. Based on the assumption, we determine the threshold for each combination by interpolating the median in the correlation graph of the evaluation results and the similarity. We select the largest one of three interpolated values of each question (Q1, Q2, and Q3). The data statistics for each combination filtered by the threshold are shown in Table~\ref{table:filter_by_thres}. 

Since the thresholds for each combination are determined differently, there are differences in the number of dialogue instances by combination. Such results suggest that the quality of multi-modal dialogue generation may vary depending on combining the text and image datasets. For example, the DailyDialog goes well with the MS-COCO but not with Flicker 30k. On the contrary, the EmpatheticDialogues goes well with the Flicker 30k but not with MS-COCO.  Thus, we must consider finding the right combination among text and image datasets in the multi-modal dialogues generation process.

\clearpage
\onecolumn
\section{Human Evaluation System}
\label{humanevalsys}

\begin{figure*}[hbt]
    \centering
    \includegraphics[width=\columnwidth]{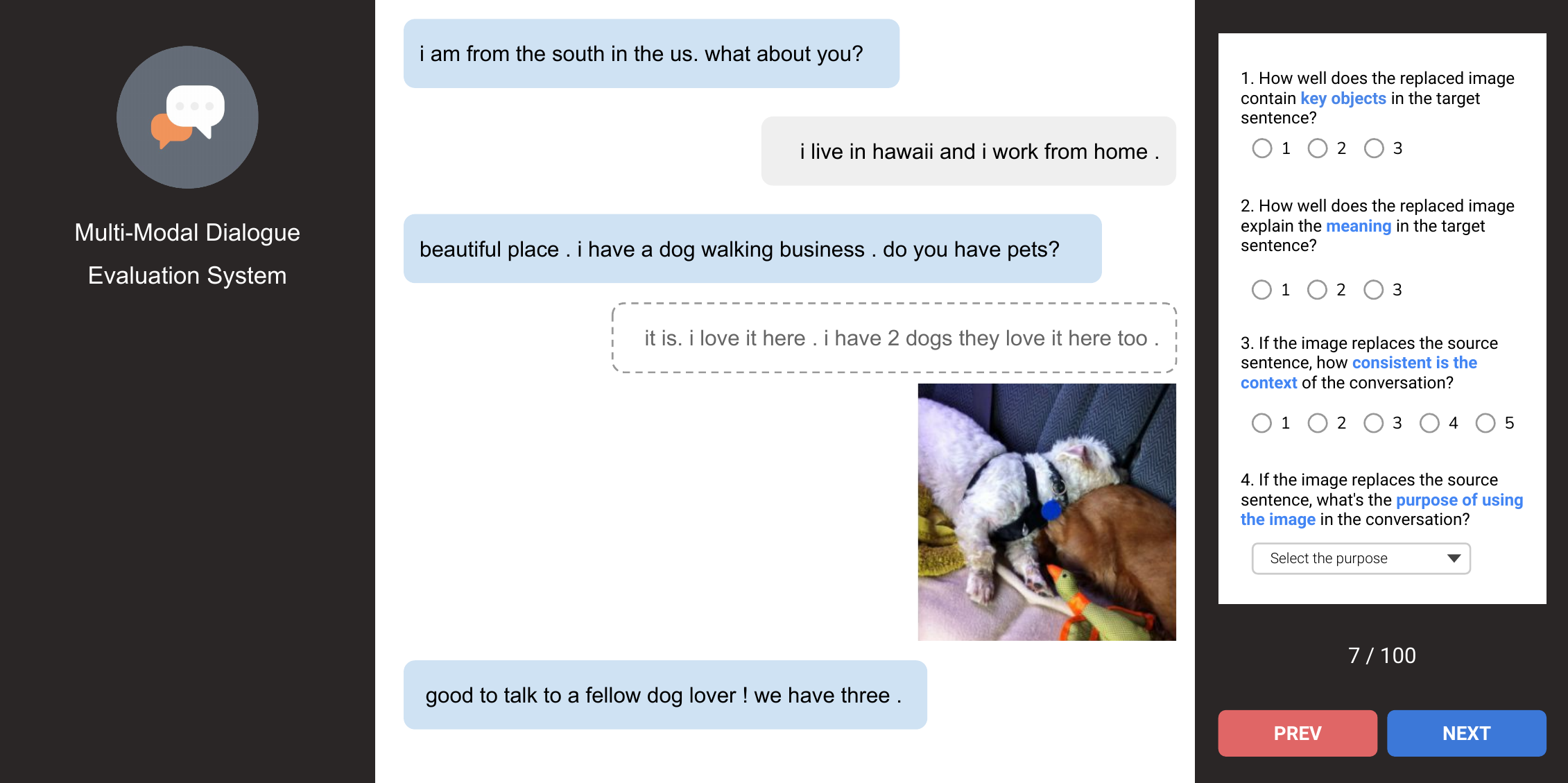}
    \caption{Human evaluation system for testing our multi-modal dialogue dataset.}
    \label{fig:evaluation_data}
\end{figure*}

\begin{figure*}[hbt]
    \centering
    \includegraphics[width=\columnwidth]{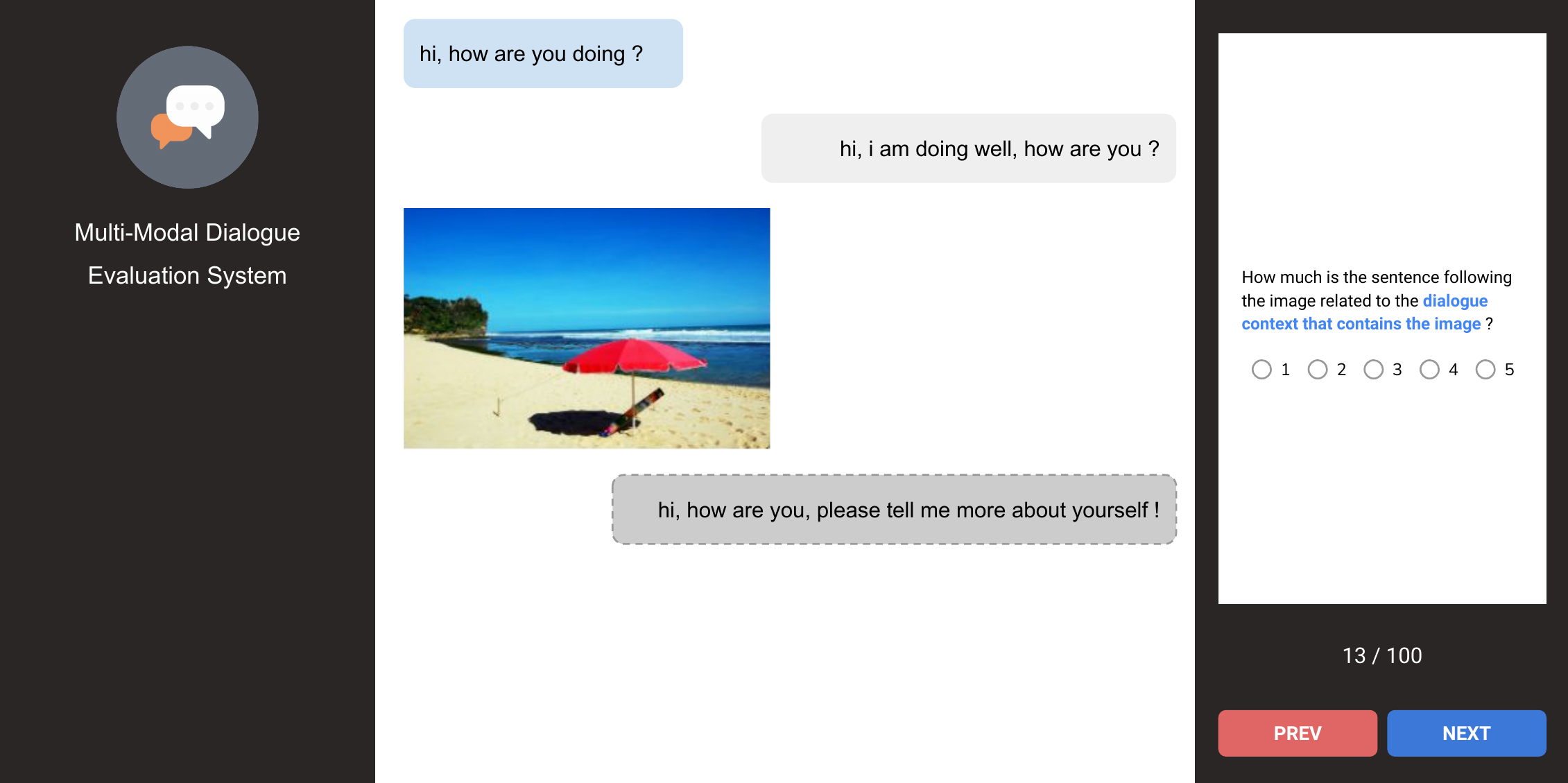}
    \caption{Human evaluation system for testing two dialogue sentence prediction tasks using our retrieval models.}
    \label{fig:evaluation_prediction}
\end{figure*}

In this section, we introduce the human evaluation system. We develop the system using a JavaScript library called ReactJS. Fig.~\ref{fig:evaluation_data} shows the implemented system for evaluating our multi-modal dialogue dataset. In this system, we ask users to evaluate a total of 100 dialog instances and answer three or four questions per instance. In addition to three questions described in Section~\ref{datacreation}, Q4\footnote{Q4: If the image replaces the source sentence, what is the purpose of using the image in the conversation?} is added depending on the purpose of use. Fig.~\ref{fig:evaluation_prediction} shows the system for evaluating the performance of a retrieval model that performs dialog sentence prediction tasks. Similarly, we also ask users to evaluate a total of 100 dialog instances and answer one question per instance.

\clearpage
\onecolumn
\section{Best Model Search}
\label{bestmodel}

\begin{table*}[hbt]
\centering
    {\small
    {\tabulinesep=0.6mm
    \begin{tabu}{llll|ccc}
    Model & Fusion Module & Image Encoder & Text Encoder & R@1 & R@5 & Mean Rank\\
    \hline
    $IR Baseline$ & n/a & n/a & n/a & 21.62 & 49.49 & 30.04\\
    \hline
    $Retrieval Model_{Att}$ & Attention & Unfreeze & Freeze & 11.74 & 39.13 & 15.73\\
    $Retrieval Model_{Sum}$ & Sum & Unfreeze & Freeze & 9.95 & 35.13 & 15.73\\
    \hline
    $Retrieval Model_{Att}$ & Attention & Unfreeze & Unfreeze & 43.51 & 80.55 & 4.13\\
    $Retrieval Model_{Sum}$ & Sum & Unfreeze & Unfreeze & 48.19 & 84.21 & 3.66\\
    \hline
    $Retrieval Model_{Att}$ & Attention & Freeze & Unfreeze & 48.41 & 85.97 & 3.40\\
    $Retrieval Model_{Sum}$ & Sum & Freeze & Unfreeze & \textbf{50.35} & \textbf{86.64} & \textbf{3.11}\\
    \end{tabu}
    }}
\caption{\label{table:bestcurrent}Comparison tests of the current dialogue prediction task on the multi-modal dialogue dataset. We compare different module variations and training strategies for our retrieval models.}
\end{table*}

\begin{table*}[hbt]
\centering
    {\small
    {\tabulinesep=0.6mm
    \begin{tabu}{llll|ccc}
    Model & Fusion Module & Image Encoder & Text Encoder & R@1 & R@5 & Mean Rank\\
    \hline
    $IR Baseline$ & n/a & n/a & n/a & 8.13 & 21.07 & 29.41\\
    \hline
    $Retrieval Model_{Att}$ & Attention & Unfreeze & Freeze & 2.04 & 9.50 & 40.99\\
    $Retrieval Model_{Sum}$ & Sum & Unfreeze & Freeze & 3.08 & 12.46 & 36.36\\
    \hline
    $Retrieval Model_{Att}$ & Attention & Unfreeze & Unfreeze & 4.09 & 15.95 & 32.07\\
    $Retrieval Model_{Sum}$ & Sum & Unfreeze & Unfreeze & 13.38 & 33.93 & 21.10\\
    \hline
    $Retrieval Model_{Att}$ & Attention & Freeze & Unfreeze & 10.02 & 28.49 & 23.71\\
    $Retrieval Model_{Sum}$ & Sum & Freeze & Unfreeze & \textbf{14.38} & \textbf{36.10} & \textbf{20.58}\\
    \end{tabu}
    }}
\caption{\label{table:bestnext}Comparison tests of the next dialogue prediction task on the multi-modal dialogue dataset. We compare different module variations and training strategies for our retrieval models.}
\end{table*}

We compare different module options of our model. Each encoder has two options: whether to freeze or not during training, and the fusion module has two options: summation, and the attention-based transformer encoder. For final image-context fused representation, context and image representations are added in the summation fusion method, while two representations are concatenated, and then fed into the attention-based two-layer transformer encoder in the attention-based method. By this comparison, we decide to freeze only the image encoder and use the summation fusion method for both current and next dialogue prediction tasks.

We additionally show the results of an information retrieval baseline, which retrieves target dialogue using the tf-idf method between candidate dialogues and the caption of an image followed by dialogue context. As shown in Tables~\ref{table:bestcurrent} and~\ref{table:bestnext}, our retrieval model significantly outperforms the information retrieval baseline, indicating that comprehensive understanding of context and images is helpful in multi-modal dialogues.

Our implementation uses an NVIDA TITAN RTX GPU for training, and training each epoch takes about 15 minutes. Our retrieval model using the summation fusion method has 204M parameters, while that using the attention-based fusion method has 254M parameters.
\clearpage
\onecolumn
\section{Multi-Modal Dialogue Dataset Example}
\label{datasetexample}

\begin{figure*}[hbt]
\centering
\begin{tabular}{cc}
     \includegraphics[width=0.5\textwidth]{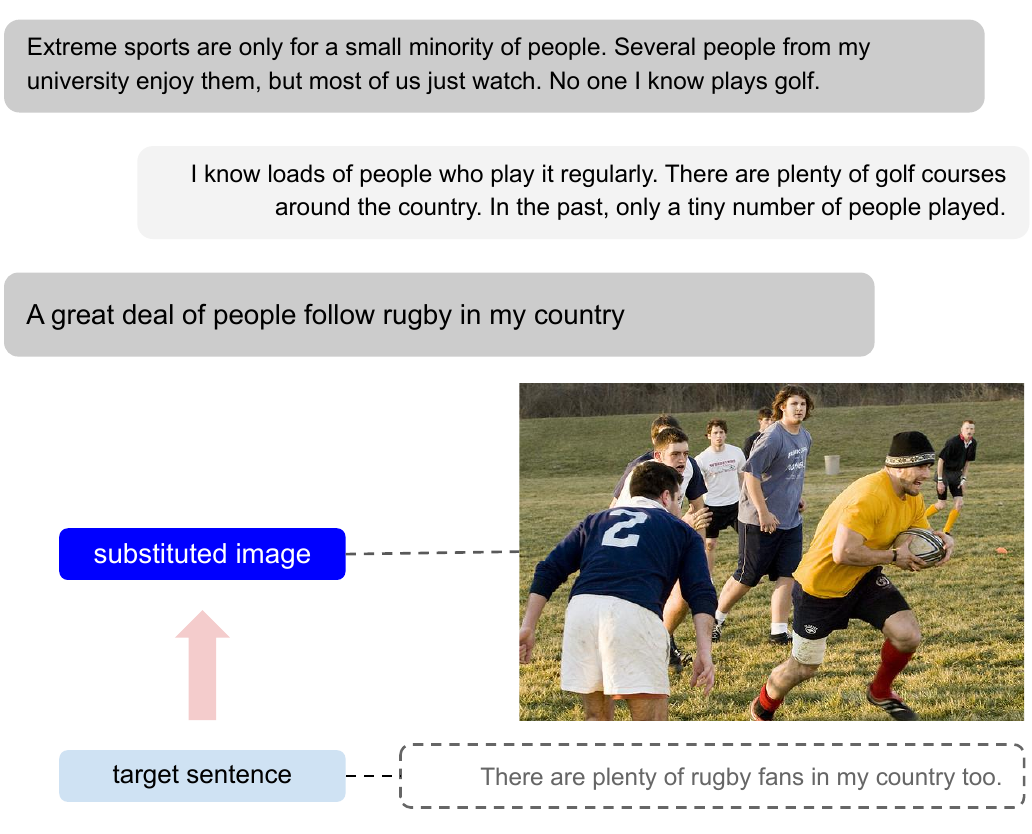}
     \includegraphics[width=0.5\textwidth]{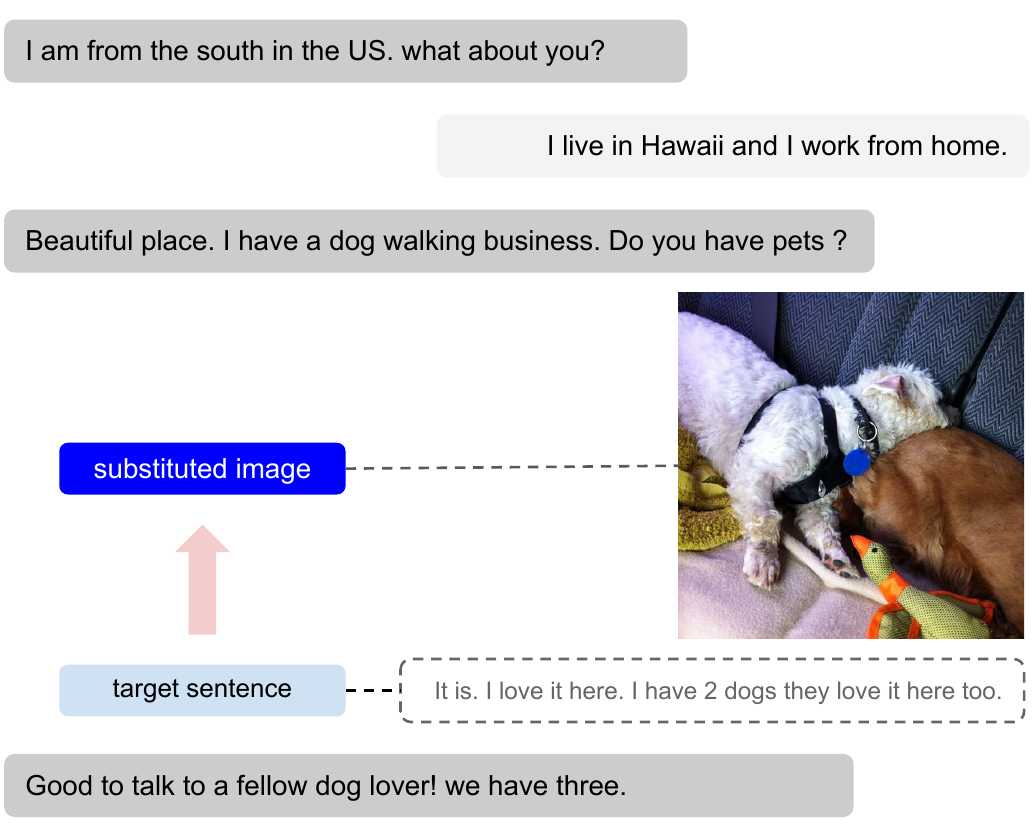}
\end{tabular}
\caption{Our multi-modal dialogue dataset examples}
\label{fig:example_mmdataset}
\end{figure*}

For easy understanding of our dataset, we provide two additional examples of the multi-modal dialogue dataset in Fig.~\ref{fig:example_mmdataset}.

\section{Selected Example of Current Dialogue Prediction Task}
\label{taskexample}

\begin{figure*}[hbt]
    \centering
    \includegraphics[width=0.7\columnwidth]{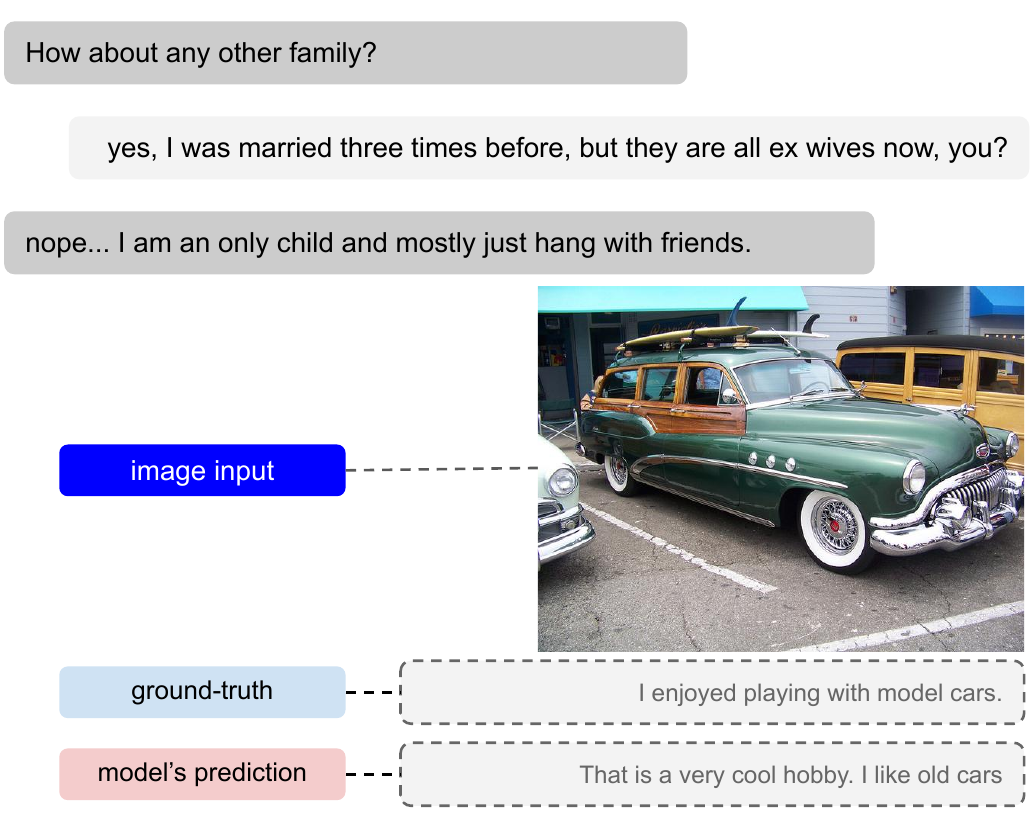}
    \caption{Ground-truth and dialogue sentence prediction example by our retrieval model used in the current turn prediction task.}
    \label{fig:example_prediction}
\end{figure*}

Fig.~\ref{fig:example_prediction} shows a reasonable example of a retrieved dialogue sentence by the retrieval model used in the current turn prediction task. Even if the model does not predict the ground-truth sentence, it can predict a plausible dialogue sentence.
\end{document}